\documentclass{article}

\usepackage{microtype}
\usepackage{graphicx}
\usepackage{subfigure}
\usepackage{booktabs} % for professional tables
\usepackage{pifont}
\usepackage[skip=0pt]{caption}
\usepackage{balance}

\usepackage{hyperref}

\usepackage[accepted]{icml2025}

\usepackage{amsmath}
\usepackage{amssymb}
\usepackage{mathtools}
\usepackage{amsthm}

\usepackage[capitalize,noabbrev]{cleveref}

\theoremstyle{plain}

\theoremstyle{definition}

\theoremstyle{remark}

\usepackage[textsize=tiny]{todonotes}

\newcommand{\algo}{\textit{LegoLink}}
\newcommand{\system}{\textsc{Epic}}
\newcommand{\baseline}{CacheBlend}

\begin{document}

\twocolumn[
\icmltitle{\system: Efficient Position-Independent Caching for Serving Large Language Models}



\icmlsetsymbol{equal}{*}
\icmlsetsymbol{intern}{$*$}

\begin{icmlauthorlist}
\icmlauthor{Junhao Hu$^\dag$}{PKU}
\icmlauthor{Wenrui Huang}{NJU}
\icmlauthor{Weidong Wang}{NJU}
\icmlauthor{Haoyi Wang}{PKU}
\icmlauthor{Tiancheng Hu}{PKU}
\icmlauthor{Qin Zhang}{HW}
\icmlauthor{Hao Feng}{HW}
\icmlauthor{Xusheng Chen}{HW}
\icmlauthor{Yizhou Shan}{HW}
\icmlauthor{Tao Xie}{PKU+}
% \icmlauthor{Firstname2 Lastname2}{equal,yyy,comp}
% \icmlauthor{Firstname3 Lastname3}{comp}
\end{icmlauthorlist}

\icmlaffiliation{PKU}{SCS, Peking University; Key Lab of HCST (PKU), MOE, China}
\icmlaffiliation{PKU+}{Key Lab of HCST (PKU), MOE; SCS, Peking University, China}
\icmlaffiliation{HW}{Huawei Cloud, Shanghai, China}
\icmlaffiliation{NJU}{School of Computer Science, Nanjing University, Nanjing, China}

\icmlcorrespondingauthor{Tao Xie}{taoxie@pku.edu.cn}
\icmlcorrespondingauthor{Yizhou Shan}{shanyizhou@huawei.com}

\icmlkeywords{Machine Learning System, KV Cache, Attention Sparsity, Large Language Models}

\vskip 0.3in
]



\printAffiliationsAndNotice{$^\dag$This work was completed during his internship at Huawei Cloud.}

% \renewcommand{\thefootnote}{\fnsymbol{footnote}}
% \footnotetext[2]{Work was completed during his internship at Huawei Cloud.}
% \renewcommand{\thefootnote}{\arabic{footnote}}

\begin{abstract}
Large Language Models (LLMs) show great capabilities in a wide range of applications, but serving them efficiently becomes increasingly challenging as requests (prompts) become more complex. Context caching improves serving performance by reusing Key-Value (KV) vectors, the intermediate representations of tokens that are repeated across requests. However, existing context caching requires exact prefix matches across requests, limiting reuse cases in settings such as few-shot learning and retrieval-augmented generation, where immutable content (e.g., documents) remains unchanged across requests but is preceded by varying prefixes.
Position-Independent Caching (PIC) addresses this issue by enabling modular reuse of the KV vectors regardless of prefixes. We formalize PIC and advance prior work by introducing \system, a serving system incorporating our new \algo\ algorithm, which mitigates the inappropriate  ``attention sink'' effect at every document beginning, to maintain accuracy with minimal computation. Experiments show that \system\ achieves up to $8\times$ improvements in Time-To-First-Token (TTFT) and $7\times$ throughput gains over existing systems, with negligible or no accuracy loss.
\end{abstract}

\section{Introduction}
\label{sec-intro}

Large Language Models (LLMs) are now fundamental to various emerging applications such as question answering, chatbots, education, and medicine~\cite{zhou2024survey}.
Users interact with LLMs by submitting requests, or prompts that consist of text-like tokens.
As LLMs’ capabilities continue to grow, their usage has shifted from simple dialogues to more complex tasks, such as multi-document question answering, few-shot learning, and tool use.
These tasks typically involve long prompts comprising relatively immutable token chunks (compared to mutable user instructions) such as system messages, few-shot examples, and documents. Notably, such immutable chunks are frequently repeated across requests.

\begin{figure}[t]
    \centering
    \includegraphics[width=1\linewidth]{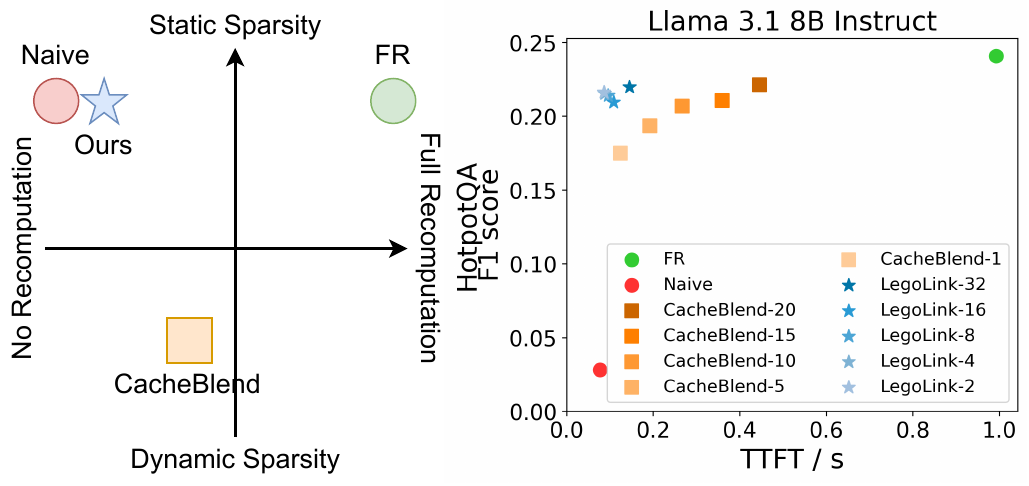}
    \caption{Left: Design space of position-independent context caching. Right: The x-axis shows the computation overhead or TTFT, while the y-axis shows accuracy. Different shades of the same color indicate variants of the same algorithm.}
    \label{fig-intro-positioning}
    \vskip -0.1in
\end{figure}

% \textit{Naive} concatenates the pre-generated KV cache of different token chunks without recomputing any tokens, achieving the lowest time-to-first-token (TTFT) and the lowest accuracy. \textit{Fully Recompute (FR)} recomputes all tokens. \baseline-15 select (at runtime) around $15\%$ tokens to recompute, utilizing dynamic attention sparsity. \algorithm-20 (this work) select (at offline) $20$ tokens on each chunk boundary to recompute, utilizing static attention sparsity. (b) Our algorithm outperforms others at both accuracy and \textit{TTFT} while running Llama3.1-8B with \textit{HotpotQA} dataset (More results later).

% \Description[]{} % Needed to avoid warning
%Context caching. Raw tokens are tokens that an LLM needs to compute to get KV caches. Cached tokens are tokens whose KV caches are already cached. (a) Full KV recomputation without context caching introduces redundant computation of Doc1. (b) Context caching eliminates the preceding redundancy by cache-and-reuse. (c) Context caching does not support direct concatenations of cached KV cache. Because direct concatenations introduce accuracy degradation due to the lack of cross-attention between Doc1 and Doc2~\cite{yao2024cacheblend}.

Context Caching\footnote{Also referred to as prompt caching.} (CC) is an emerging approach that reuses Key-Value (KV)  vectors, the intermediate representations of repeated tokens in previous requests to reduce computation, and is generally categorized into two types. 
First, \textbf{prefix-based CC} matches the current request against previous ones to reuse the KV vectors of the longest common prefix. Although prefix-based CC remains the dominant approach in existing systems~\cite{kimi-context-caching, gemini-context-caching, zheng2023sglang, kwon2023vllm}, it requires exact prefix matches across requests, limiting reuse cases in settings such as few-shot learning and Retrieval-Augmented Generation (RAG), where immutable chunks (e.g., documents) remain unchanged across requests but are preceded by varying prefixes. Second, \textbf{Position-Independent Caching (PIC)} (Figure~\ref{fig-background-compiling-analogy} (b)) extends prefix-based CC, enabling modular reuse of the KV vectors of immutable tokens, regardless of their prefixes~\cite{yao2024cacheblend}. Although PIC significantly increases reuse opportunities (Figure~\ref{fig-eval-e2e}), it deviates from standard attention mechanisms, resulting in potential accuracy degradation; thus, ensuring accurate recovery becomes its main challenge.

To tackle the challenge of PIC, we formalize its usage within a two-step framework analogous to compilation and linking (Figure~\ref{fig-background-compiling-analogy}). First, the \textbf{compile} step involves submitting individual immutable chunks to the LLM to generate and store their respective KV vectors. Second, the \textbf{link} step retrieves and concatenates cached KV vectors and recomputes a subset of KV vectors to maintain accuracy.

To the best of our knowledge, \baseline~\cite{yao2024cacheblend} is the first\footnote{Another related approach called PromptCache~\cite{gim2024prompt} is not of the PIC type.} work that fits into our PIC framework (Figure~\ref{fig-intro-positioning}), and has two major limitations.
First, the time and resource complexity of the recomputation in the link step are the same as the original attention mechanism---$O(N^2)$, where $N$ is the number of tokens in the given prompt. Figure~\ref{fig-intro-positioning} shows that, although \baseline-15 dynamically selects $15\%$ of tokens for recomputation, for very long prompts, common in many applications today, this $O(15\%N^2)$ complexity remains slow and prone to out-of-memory (OOM) errors (Figure~\ref{fig-eval-long-context}).
Second, \baseline\ relies on dynamic attention sparsity, which incurs heavy runtime overhead in addition to the $O(N^2)$ recomputation. Figure~\ref{fig-eval-breakdown} shows that the runtime overhead of \baseline\ takes around 16.3\% to 63.56\% of Time-To-First-Token (TTFT).

To overcome the limitations of \baseline, we develop \system\ (Efficient Position-Independent Caching), a serving system that incorporates our simple but effective algorithm named \algo\ with two characteristics.
First, \algo\ reduces recomputation complexity to $O(kN) \sim O(N)$, where $k \ll N$ and increases with the number of immutable chunks instead of $N$. As described in Section~\ref{sec-eval}, $k$ could potentially become zero. Second, \algo\ relies on static attention sparsity, which selects the tokens to recompute beforehand, further improving performance. The static token selection is based on our key insight: the initial tokens of each immutable chunk disproportionately absorb attention, impeding subsequent tokens from attending to relevant context---a phenomenon known as ``attention sink''~\cite{xiao2023sink}. \algo\ recomputes $k$ ($k \le 32$) initial tokens of each chunk (except the first chunk), allowing these tokens to recognize their non-initial positions and crippling their attention-sink ability.

We implement the \system\ serving system with the \algo\ algorithm based on one of the most widely used inference frameworks, vLLM~\cite{kwon2023vllm}. We evaluate \system\ against the state-of-the-art \baseline~\cite{yao2024cacheblend} system, across six tasks with distinct characteristics and three model architectures with diverse training recipes. Compared to \baseline, \system\ achieves up to a $3\times$ improvement in TTFT with accuracy losses limited to within $7\%$ (Figure~\ref{fig-intro-positioning}) when serving single requests. Furthermore, \system\ provides up to an $8\times$ reduction in TTFT and a $7\times$ increase in throughput when serving multiple requests under varying rates. The code is available at: \href{https://github.com/DerekHJH/epic}{https://github.com/DerekHJH/epic}.

In summary, this paper makes three major contributions:
\begin{itemize}
    \item We formalize the PIC usage into a two-step framework, within which we consolidate existing literature and highlight potential directions for future research.
    \item We provide a detailed analysis of existing algorithms, based on which we propose a new \algo\ algorithm, which reduces up to $3\times$ \textit{TTFT} while keeping accuracy losses limited to within $7\%$, compared to the state-of-the-art \baseline\ system.
    \item We implement the  \system\ serving system by incorporating OpenAI-compatible context caching APIs, a KV store, and \algo. \system\ reduces up to $8\times$ \textit{TTFT} and increases up to $7\times$ throughput when serving multiple requests under varying rates. 
\end{itemize}

\section{Background and Motivation}
\label{sec-background}

This section provides a primer on transformers, context caching, and its variant, Position-Independent Caching (PIC), along with a review of an existing PIC algorithm.

\subsection{Autoregressive Generation and KV Cache}
The generation process of Large Language Models (LLMs) consists of two distinct stages: the prefill stage and the decode stage.
In the \textbf{prefill} stage, the model processes a sequence of prompt tokens all at once. It computes the Key (K) and Value (V) vectors for all prompt tokens, stores these vectors in the KV cache, and generates the first output token to initiate the decode stage. The time required to generate the first token is referred to as the Time-To-First-Token (TTFT). The prefill stage is primarily compute-bound, as it involves processing multiple tokens in parallel.
In the \textbf{decode} stage, the model iteratively processes each newly generated token. It computes the KV vectors for the new token, appends these vectors to the KV cache, and generates the next token. This process repeats until a specified stopping criterion is met. Unlike the prefill stage, the decode stage is memory-bound as it computes little compared to the amount of memory access.
 
\subsection{Context Caching} 

LLMs' usage has shifted from simple dialogues to more complex tasks, such as multi-document question answering, few-shot learning, and tool use.
These tasks typically involve long prompts comprising relatively immutable token chunks (compared to mutable user instructions/queries) such as system messages, few-shot examples, and documents. Notably, such immutable chunks are frequently repeated across requests. Context Caching (CC), also referred to as prompt caching, is an emerging approach that reuses the KV vectors of repeated tokens in previous requests~\cite{hu2024memserve, zheng2023sglang, liu2023cachegen, kwon2023vllm}, speeding up the prefill stage and reducing TTFT. Context caching can be categorized into two types: prefix-based caching and Positional-Independent Caching (PIC).

\textbf{Prefix-based caching}, implemented in nearly all existing context caching systems~\cite{kwon2023vllm, zheng2023sglang, gim2024prompt}, matches the current request against previous ones to reuse the KV vectors of the longest common prefix. This approach, however, requires an exact prefix match, as each token’s KV vector depends on all preceding tokens and their absolute position IDs in the prompt. Consequently, even minor differences in the prefix invalidate the KV vectors of otherwise immutable chunks, requiring full recomputation. This constraint significantly limits reuse opportunities~\cite{lmcache-github, yao2024cacheblend}, especially in scenarios such as multi-document question answering or Retrieval-Augmented Generation (RAG), where immutable chunks (e.g., documents) remain unchanged across requests but are preceded by varying prefixes.

\begin{figure}[t]
    \centering
    \includegraphics[width=1\linewidth]{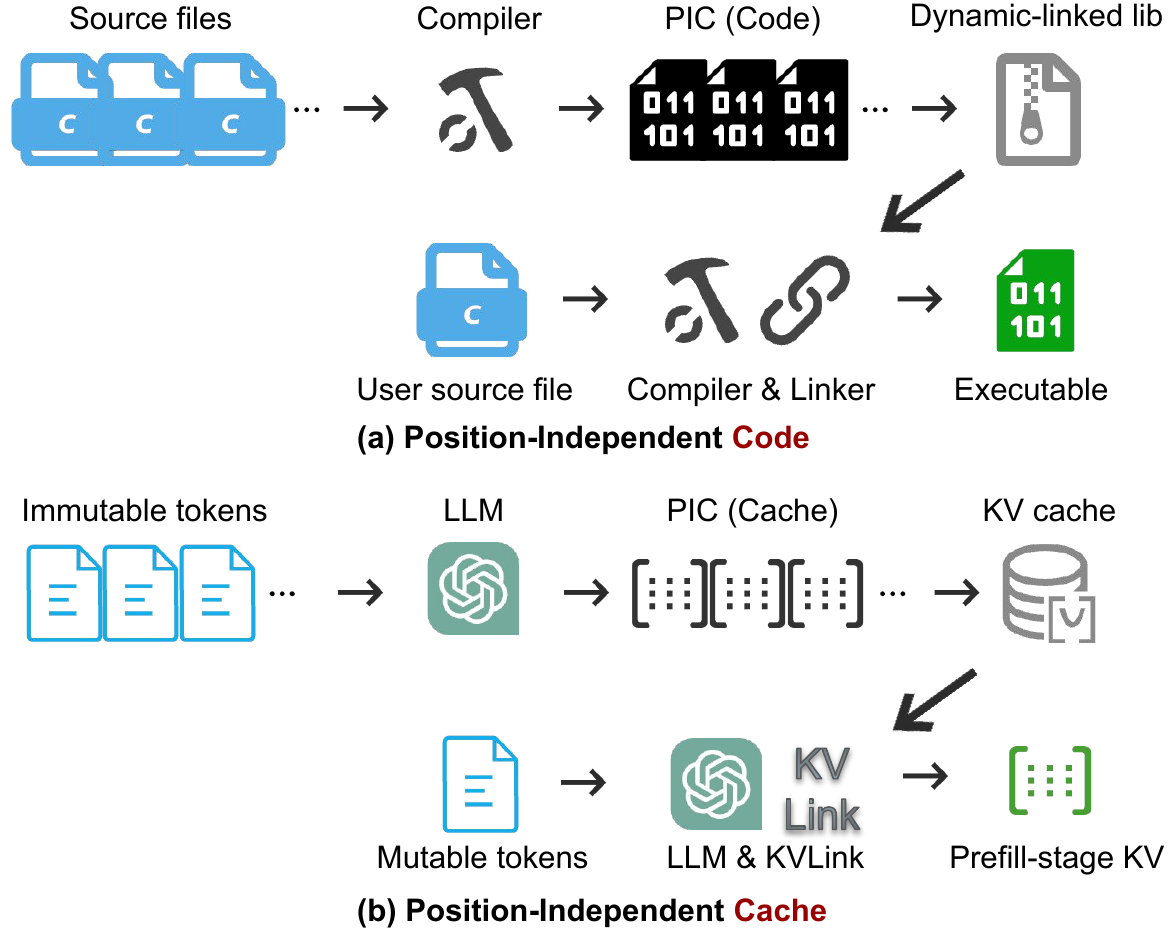}
    \caption{An analogy between position-independent code and position-independent cache.}
    \label{fig-background-compiling-analogy}
\end{figure}

\textbf{Position-Independent Caching (PIC)}, inspired by the classical position-independent code\footnote{https://en.wikipedia.org/wiki/Position-independent\_code} that can be executed at any memory address~\cite{hu2023pcrml}, enables modular reuse of the KV vectors of immutable tokens~\cite{yao2024cacheblend}, regardless of their prefix (Figure~\ref{fig-background-compiling-analogy}). PIC significantly increases reuse opportunities (Figure~\ref{fig-eval-e2e}), but it deviates from standard attention mechanisms, resulting in potential accuracy degradation; thus, ensuring accurate recovery becomes its main challenge.

We formalize PIC usage within a two-step framework. First, the \textbf{compile} step involves submitting individual immutable chunks to the LLM to generate and store their respective KV vectors. In this step, each chunk is encoded with position IDs starting from zero, and the LLM performs only the prefill stage---generating KV vectors without any prefix or further token generation. This process is analogous to compiling C source files into position-independent relocatable code. The resulting KV vectors are stored in a cache, conceptually similar to packaging object code into a dynamically linked library. Second, the \textbf{link} step retrieves and concatenates cached KV vectors and recomputes a subset of KV vectors to mitigate accuracy degradation due to deviations from the standard attention mechanism. This recomputation involves both cached tokens and uncached tokens, such as user instruction/query tokens, which are computed for the first time\footnote{Strictly speaking, uncached tokens are not recomputed, but for simplicity, we use ``recomputation'' to refer to both cases.}. This process is analogous to linking dynamically linked libraries with source code to produce an executable.
    
\subsection{Existing Algorithms for PIC}

\baseline~\cite{yao2024cacheblend} is the first PIC algorithm focusing on the link step, but two simpler algorithms also deserve attention, although they are too rudimentary to be classified as proper algorithms. First, \textit{Naive} reuses cached KV vectors directly in the link step, without any recomputation (Figure~\ref{fig-algo}, first row below the dashed line). This algorithm incurs zero linking overhead but leads to substantial accuracy degradation (Figure~\ref{fig-eval-e2e}) due to violations of the attention mechanism---the PIC challenge. Second, Fully Recompute (\textit{FR}) recomputes all KV vectors in the link step (Figure~\ref{fig-algo}, second row below the dashed line). This algorithm preserves the standard attention mechanism and achieves the highest accuracy, but it eliminates the efficiency benefits of caching, resulting in the highest linking overhead (Figure~\ref{fig-eval-e2e}).

To strike a balance between accuracy and linking overhead, \baseline~\cite{yao2024cacheblend} works as follows.
First, it retrieves the needed KV vectors and concatenates them to obtain KV\_old. Second, it recomputes all KV vectors in the first layer of the LLM, generating KV\_new\_1. Third, it compares the attention maps produced by KV\_old\_1 and KV\_new\_1, selecting the 15\% of tokens that exhibit the most discrepancy. Fourth, it recomputes only these 15\% of tokens in all subsequent layers\footnote{The recomputation rate might decrease in deeper layers.}. Only 15\% of tokens are necessary because attention exhibits \textbf{sparsity}---only a small subset of tokens significantly influence attention computation.

However, \baseline\ has two limitations. First, the time and resource complexities of the recomputation in the link step are the same as the original attention mechanism---$O(N^2)$, where $N$ is the number of tokens in the prompt. Figure~\ref{fig-intro-positioning} shows that, although \baseline-15 dynamically selects $15\%$ of tokens for recomputation, for very long prompts, common in many applications today, this $O(15\%N^2)$ complexity remains slow and prone to out-of-memory (OOM) errors (Figure~\ref{fig-eval-long-context}).
Second, \baseline\ relies on dynamic attention sparsity---recomputing all KV vectors in the first layer; this recomputation incurs heavy runtime overhead in addition to the $O(N^2)$ recomputation. Figure~\ref{fig-eval-breakdown} shows that the runtime overhead of \baseline\ takes around 16.3\% to 63.56\% of Time-To-First-Token (TTFT).

\section{System Overview}
\label{sec-overview}

\begin{figure}[t]
    \centering
    \includegraphics[width=\linewidth]{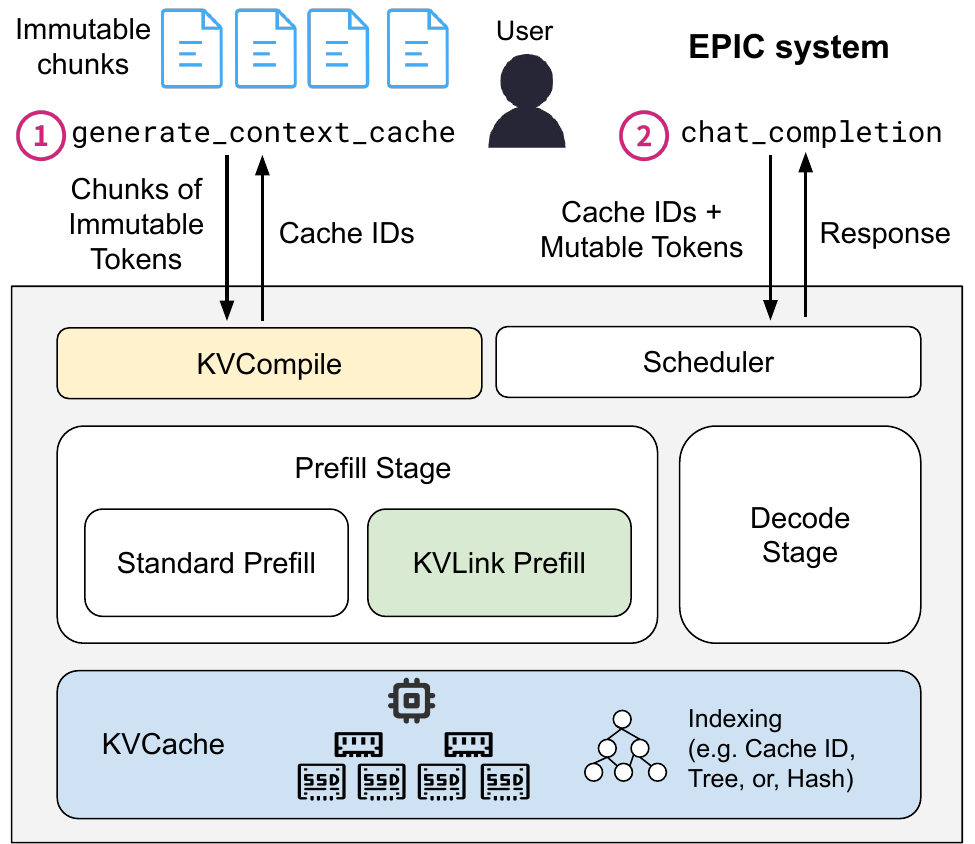}
    \caption{The architecture of \system\ serving system.}
    \label{fig-overview-system}
\end{figure}

To support PIC, we develop \system\ (Efficient Position-Independent Caching), a serving system whose workflow aligns with the PIC framework described in Section~\ref{sec-background} and consists of two main steps (Figure~\ref{fig-overview-system}). First, in the compile step, \ding{172} users submit immutable chunks via the context caching API. The \texttt{KVCompile} component processes each chunk using a standard prefill pass to generate KV vectors, which are then stored in the \texttt{KVCache}. For each chunk, \texttt{KVCompile} returns a unique cache ID, which users can later reference to enable KV reuse. Second, in the link step, \ding{173} users submit requests containing mutable tokens (e.g., new instructions/queries) along with cache IDs (if any), using an extended chat completion API. The \texttt{Scheduler} component handles these requests by initiating a \texttt{KVLink} prefill. \texttt{KVLink} retrieves the relevant KV vectors from \texttt{KVCache} using the provided cache IDs, concatenates them, recomputes a subset of KV vectors to ensure correctness, and proceeds to the decode stage for subsequent token generation. The final response is then returned to users. At the core of \texttt{KVLink} is the \algo\ algorithm, which we detail in Section~\ref{sec-algo}.

\textbf{Discussion of implicit vs. explicit caching.} Many existing systems~\cite{kwon2023vllm, zheng2023sglang, hu2024memserve} adopt an implicit caching paradigm, where the system automatically manages cache generation, storage, and reuse through internal mechanisms such as hash tables or radix trees. In contrast, \system\ adopts an explicit caching paradigm, where users manage cache generation and reuse via explicit APIs that expose cache IDs. This paradigm, also used by systems such as Google Gemini and Mooncake~\cite{kimi-context-caching, gemini-context-caching}, reduces indexing overhead and provides greater user control over cache management---particularly beneficial in RAG scenarios.

\section{Algorithm Design}
\label{sec-algo}

\begin{figure*}[thbp]
    \centering
    \includegraphics[width=\linewidth]{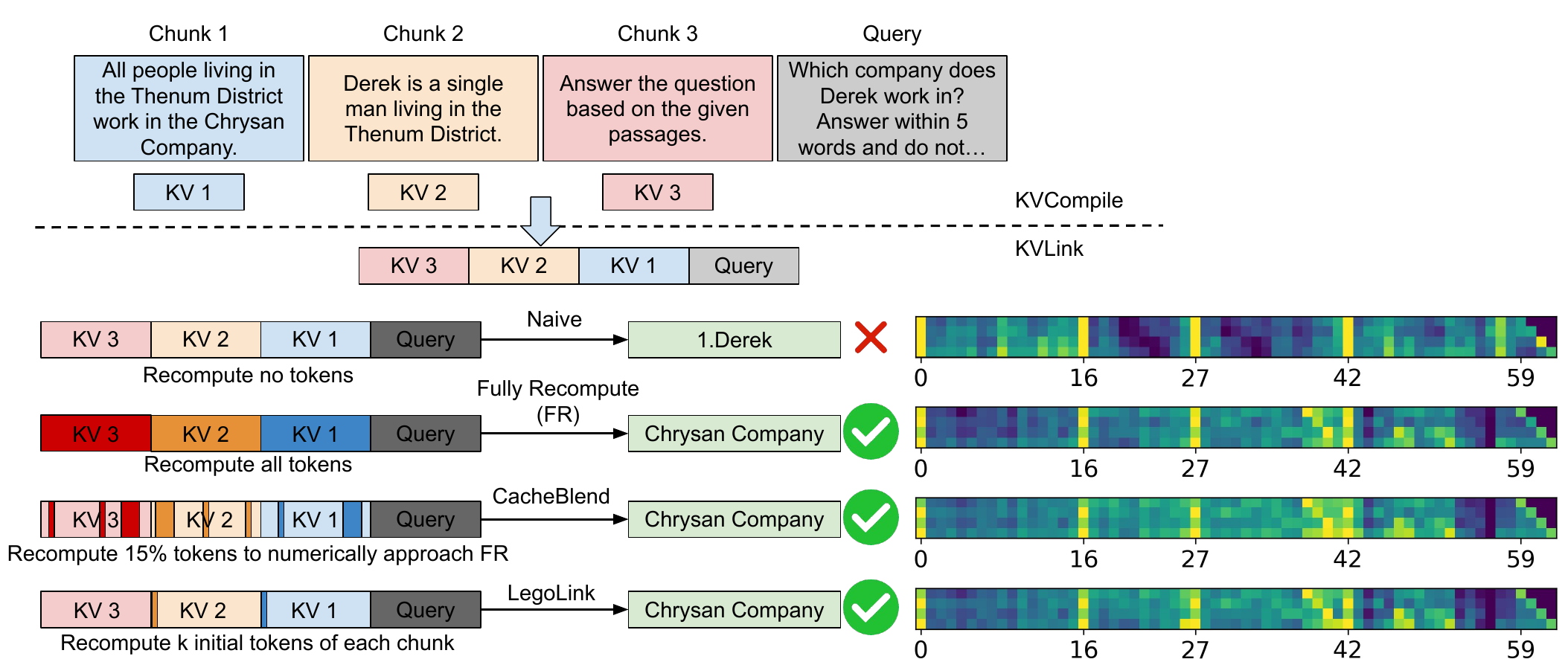}
    \caption{Comparison of PIC Algorithms. The area above the dashed line corresponds to the compile step, while the area below corresponds to the link step. \texttt{KVLink} recomputes a subset of tokens, highlighted in dark colors. Four algorithms include \textit{Naive}, \textit{Fully Recompute (FR)}, \baseline, and \algo. The bottom right visualizes attention maps (layer 5, head 5 of Llama 3.1 8B) for four decoded tokens. The x-axis marks the position ID of the first token of each chunk. To highlight the differences between attention maps, we normalize the $QK^T$ results to the [0, 1] range using min-max scaling instead of Softmax.}
    \label{fig-algo}
    
\end{figure*}

In this section, we analyze existing algorithms and propose a new algorithm \algo\ based on the analysis results.

\subsection{Analysis of Existing Algorithms}

We analyze the attention map of \textit{Naive}, \textit{FR}, and \baseline\ algorithms (Section~\ref{sec-background}), shown in the bottom right of Figure~\ref{fig-algo}. First, in the \textit{Naive}'s attention map, most attention scores concentrate on the initial tokens of each chunk, evident from the four bright vertical lines along the x-axis. This pattern arises because \system\ independently compiles each chunk with position IDs starting from zero. As a result, the initial tokens disproportionately absorb attention---a phenomenon known as ``attention sink''~\cite{xiao2023sink}---which prevents subsequent tokens from attending to the answer ``Chrysan Company,'' located at the end of the third chunk (Chunk 1). Second, in the \textit{FR}'s attention map, the initial tokens of each chunk release part of their attention to more relevant positions, including ``Chrysan Company''. However, they still retain relatively strong attention scores partly because they are special begin-of-sentence tokens, such as ``\verb|<s>|'' in Llama models. Third, in the \baseline's attention map, the pattern closely resembles that of \textit{FR}, reflecting its design goal of approximating \textit{FR}'s attention map. Further analysis of \baseline's selected 15\% of tokens shows that initial tokens of each chunk are frequently included, reinforcing the importance of recomputing these tokens to improve accuracy.

\subsection{The \algo\ Algorithm}

Based on the preceding analysis, we propose \algo, which recomputes each chunk's first $k$ tokens (except the first chunk), thus linking chunks like Lego pieces. By recomputing these initial tokens, \algo\ enables them to recognize their non-initial status, thereby mitigating their tendency to dominate attention and redirecting attention to relevant positions, as shown in \algo's attention map in Figure~\ref{fig-algo}. Evaluation results in Section~\ref{sec-eval} demonstrate that \algo\ consistently preserves accuracy across a wide range of datasets and models. See \algo's details in the next paragraph.

Assuming that we have selected $k'$ ($k$ tokens from each chunk plus the user query) tokens from a total of $N$ (prompt length) tokens, we recompute them as follows. First, we obtain the embedding matrix E (with shape $(k', d)$) of the $k'$ tokens, where $d$ is the hidden size. Second, at layer $i$, we compute the new K, Q, and V matrices (each with shape $(k', d)$) for these $k'$ tokens: $Q = EW_Q$, $K = EW_K$, $V = EW_V$, where $W_Q$, $W_K$, and $W_V$ are model parameters with shape $(d, d)$\footnote{For notation simplicity, $d$ represents all possible hidden dimension sizes, which may be further divided into the number of heads and head dimensions.}. Third, we expand the K and V matrices by incorporating the cached KV vectors of the $N - k'$ unselected tokens at correct positions, forming $K_{exp}$ and $V_{exp}$ (both with shape $(N, d)$). Fourth, we compute the attention matrix A (with shape $(k', N)$) by multiplying Q (with shape $(k', d)$) with $K_{exp}^T$ (with shape $(d, N)$), allowing the $k'$ tokens to attend to all $N$ tokens:
\begin{align}
A = \text{softmax}(QK_{exp}^T\cdot \text{MASK})
\end{align}
where MASK assures that the $k'$ tokens attend to only tokens before them. Finally, we multiply A (with shape $(k', N)$) with $V_{exp}$ (with shape $(N, d)$) to obtain the output (or input to the next layer, with shape $(k', d)$): $O = AV_{exp}W_O$, where $W_O$ is a matrix with shape $(d, d)$.

In addition to preserving accuracy and simple design/implementation, \algo\ offers two key advantages over \baseline. First, \algo\ reduces recomputation complexity to $O(kN) \sim O(N)$, where $k \ll N$ and increases with the number of immutable chunks instead of $N$. As described in Section~\ref{sec-eval}, $k$ could potentially become zero. Second, \algo\ relies on static attention sparsity, which selects each chunk's first $k$ tokens to recompute beforehand, further improving performance. 
\section{Evaluation}
\label{sec-eval}

We begin by describing the experimental setup, including implementation details, datasets, models, evaluation metrics, and software/hardware environment. We then present four key evaluation results.

\subsection{Experiment Setup}

\textbf{Implementation.} We implement \system\ based on vLLM 0.4.1~\cite{kwon2023vllm}, with 2K lines of code in Python. We incorporate the four PIC algorithms presented in Figure~\ref{fig-algo}. We port CacheBlend from their public repository\footnote{https://github.com/YaoJiayi/CacheBlend. Accessed in Sep 2024.}.

\begin{figure}[t]
    \centering
    \includegraphics[width=1\linewidth]{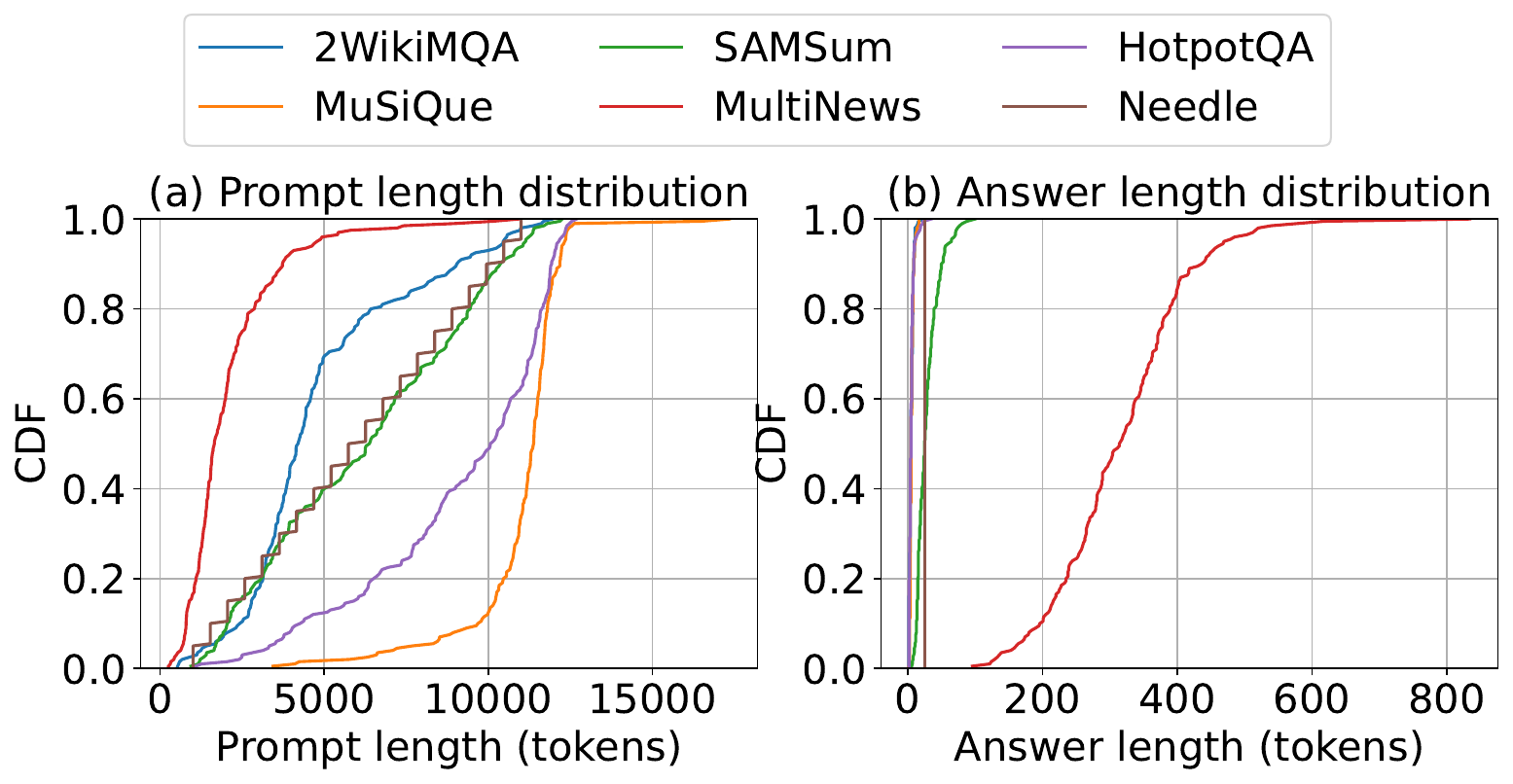}
    \caption{Prefill and decode length distribution.}
    \label{fig-eval-data-dist}
\end{figure}

\textbf{Dataset.} Following \baseline, we evaluate on four LongBench datasets~\cite{bai2024longbench}: \textit{2WikiMQA} (multi-document question answering), \textit{MuSiQue} (multi-document question answering), \textit{SAMSum} (few-shot instruction following), and \textit{MultiNews} (multi-document summarization). We also include \textit{HotpotQA} (multi-document question answering) from LongBench, which identifies the two supporting documents containing the answer, enabling fine-grained analysis. To evaluate long-context retrieval, we include the \textit{Needle in a Haystack} dataset~\cite{LLMTest_NeedleInAHaystack}, which tests the model’s ability to locate and retrieve a single inserted fact from unrelated documents of varying lengths. All datasets contain 200 test cases, with the distribution of prompt (prefill) lengths and answer (decode) lengths shown in Figure~\ref{fig-eval-data-dist}. Immutable tokens constitute approximately 95\%-99\% of the prompt, while mutable tokens are fewer than 50.

\textbf{Metrics.} We use the following three metrics to evaluate performance and model accuracy. First, \textit{Time-To-First-Token (TTFT)}~\cite{kwon2023vllm} (lower is better) is used to evaluate all datasets. This metric measures the prefill-stage time: the time from when users send a request to when users receive the first token; this time could be reduced by using context caching. Second, \textit{F1 score}~\cite{bai2024longbench} (higher is better) is used to evaluate \textit{2WikiMQA}, \textit{MuSiQue}, \textit{HotpotQA}, and \textit{needle in a haystack}. This metric measures the similarity between LLMs' output and the ground-truth answer based on their common words. Third, \textit{Rough-L score}~\cite{lin2004rouge} (higher is better) is used to evaluate \textit{SAMSum} and \textit{MultiNews}. This metric measures the similarity between LLMs' output and the ground-truth answer by calculating the length of their longest common subsequence.

\textbf{Models.} We evaluate \system\ and \algo\ using three state-of-the-art open-source LLMs: Mistral 7B Instruct~\cite{chaplot2023albert}, Llama 3.1 8B Instruct~\cite{llama3.1}, and Yi Coder 9B Chat~\cite{young2024yi}. These models represent diverse architectures and training recipes. Rather than employing quantized versions of larger models, we select smaller models to accommodate our limited GPU resources. Additionally, we do not choose models fine-tuned for the six specific task types, as the number of such models is extensive. While our chosen general-purpose base models may exhibit lower absolute accuracy on these tasks, the relative accuracy drop compared to the base model is sufficient to demonstrate the effectiveness of \system\ and \algo.

\textbf{Baselines.} We compare \algo\ with the other three recomputation algorithms in Figure~\ref{fig-algo}: \textit{FR}, \textit{Naive}, and \baseline~\cite{yao2024cacheblend}. Additionally, we evaluate different variants of \baseline, denoted as \baseline-r, where $r$ represents the ratio of tokens recomputed. Similarly, we evaluate different variants of \algo, denoted as \algo-k, where $k$ refers to each chunk's first $k$ tokens.

\textbf{Environment.} We run experiments on a single NVIDIA A100 server with one A100-80GB GPU available. The server has 128-core Intel(R) Xeon(R) Platinum 8358P CPU@2.60GHz with 2 hyperthreading and 1TB DRAM. We use Ubuntu 20.04 with Linux kernel 5.16.7 and CUDA 12.6.

\subsection{Workloads}

We construct the following two kinds of workflows out of the six datasets.

\textbf{Synchronous workload.} To evaluate the accuracy–latency trade-off without interference from concurrent requests, we process test cases sequentially, ensuring that each completes before the next begins. First, for each test case, we compile all immutable chunks to obtain their corresponding cache IDs. For the LongBench dataset (excluding SAMSum), we treat each document as a chunk. For SAMSum and Needle-in-a-Haystack, we split all immutable tokens into 512-token chunks. Second, we send a request containing the cache IDs of cached chunks along with the query to obtain the response.

\textbf{Asynchronous workload.} To evaluate the latency and throughput of \system\ under varying request rates (requests per second), we simulate a PIC scenario as follows\footnote{PIC is a relatively new approach and lacks publicly available traces or request arrival patterns. We try our best to mitigate potential bias in constructing this asynchronous workload.}. First, we select $d$ test cases from \textit{2WikiMQA} to simulate $d$ active users. As the number of users increases, a larger portion of the GPU HBM is allocated to their position-independent cache. This portion is defined as the Context Cache Ratio (CCR). Second, each user compiles all immutable chunks in the test case once and then repeatedly sends the same request (containing cache IDs of cached chunks along with the query) at a constant rate over a 40-second period. Although each user resends identical requests, this setup effectively simulates a user having different queries over the same document set, with all other context caching mechanisms, such as prefix caching, disabled. Third, we simulate request arrival times by sampling from a Poisson distribution.

\subsection{Accuracy-Latency Trade-off of \algo}

\begin{figure}[t]
\vskip -0.1in
    \centering
    \includegraphics[width=\linewidth]{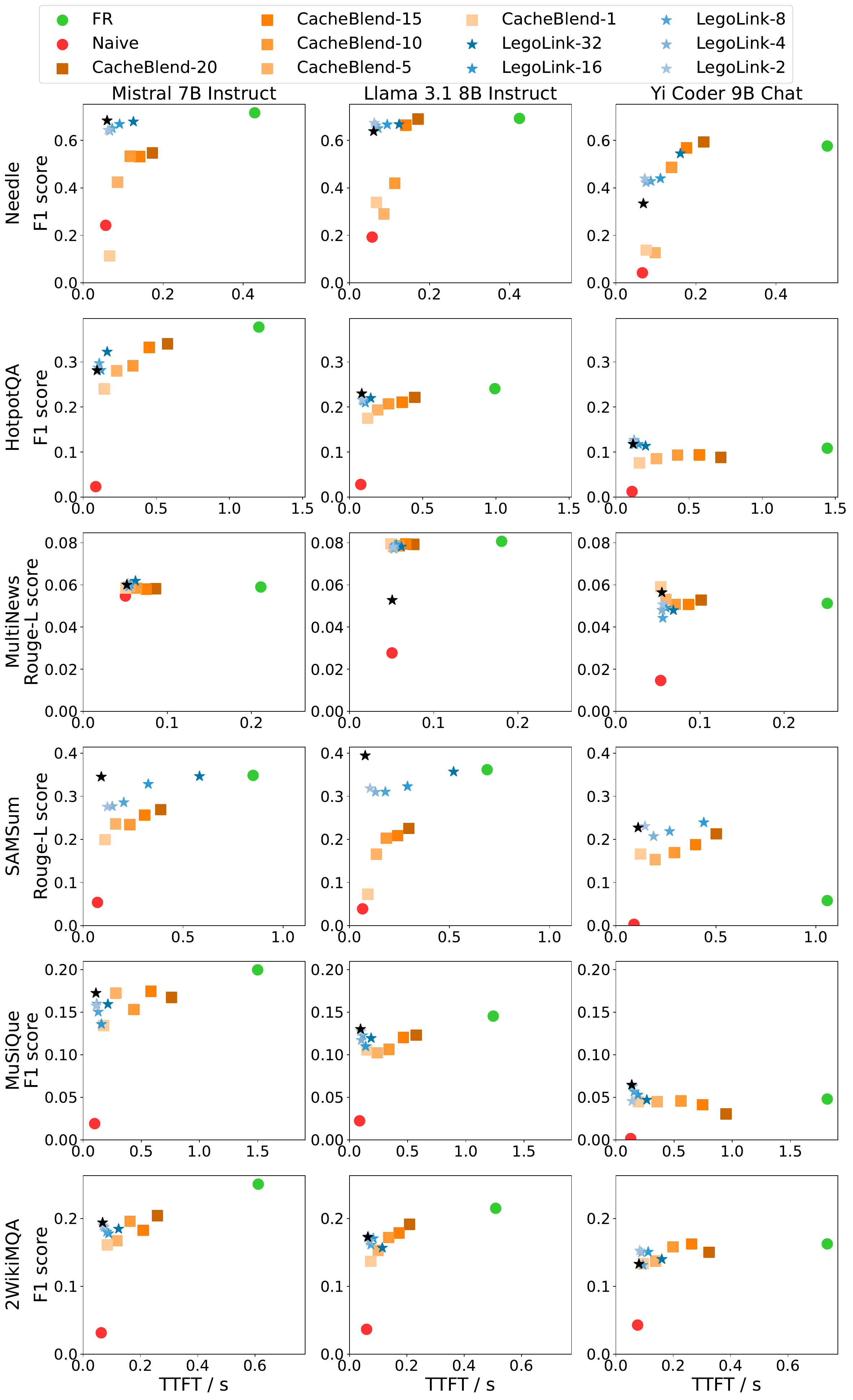}
    \caption{Accuracy vs. TTFT. Each point indicates the average \textit{TTFT} and accuracy for running synchronous workloads of one dataset (row) on one model (column) using one specific algorithm (each legend label). The $k$ in \algo-k denotes the number of recomputed initial tokens for each chunk, while the $r$ in \baseline-r represents the ratio of all recomputed tokens. The black star represents \algo-0, a zero-linking algorithm.}
    \label{fig-eval-e2e}
    \vskip -0.2in
\end{figure}
\begin{figure}[t]
    \centering
    \includegraphics[width=1\linewidth]{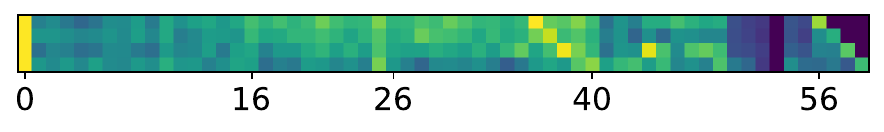}
    \caption{Attention map of \algo-0 using the example in Figure~\ref{fig-algo}.}
    \label{fig-eval-zero}
\end{figure}

Using the synchronous workload described earlier, we draw three key insights from the results in Figure~\ref{fig-eval-e2e}. First, \algo\ variants (a series of gradient blue stars) establish a new Pareto frontier, outperforming \baseline\ variants (a series of gradient orange rectangles) in most cases. Second, \algo-2 is sufficient to limit accuracy drops within 0 - 7\% and reduces up to $300\%$ \textit{TTFT}, compared to the default \baseline-15 configuration. On the contrary, \baseline-1 or \baseline-5, which recomputes a similar number of tokens as most \algo\ variants (except on \textit{SAMSum}), exhibits significant accuracy degradation---up to 80\% worse than \textit{FR}. Third, increasing the number of recomputed tokens in \algo\ yields diminishing accuracy gains. Recomputing only a small number of initial tokens suffices to restore most of the accuracy, whereas \baseline\ requires substantially more recomputation for marginal benefits.

In addition, we also have three unusual observations that warrant further explanation. First, all algorithms, including \textit{FR}, \baseline, and \algo, exhibit low accuracy in the Yi Coder model due to its poor handling of document understanding. This observation suggests that, to ensure that PIC algorithms perform optimally, robust models well suited to the task are required. Second, all approaches using all models perform poorly on the \textit{MultiNews} dataset. This phenomenon can be attributed to the inherent difficulty of summarizing long documents with small models. Third, \baseline\ and \algo\ exhibit similar \textit{TTFT} in the \textit{MultiNews} and \textit{SAMSum} datasets. Each document in these datasets is relatively short (around one hundred tokens), making the number of tokens recomputed ($k$ tokens in \algo-k) equivalent to the ratio of tokens recomputed ($r\%$ in \baseline-r).

\subsection{Algorithm Analysis}

To further understand \algo, we introduce \algo-0, a variant that shifts all linking overhead to the compile step and comprises two different PIC steps. First, \system\ prepends four dummy tokens (e.g., begin-of-sentence tokens) to each immutable chunk during compilation, then discards their corresponding KV vectors. This removal eliminates ``attention sink'' tokens in advance, preventing them from interfering with subsequent attention computations. Second, in the link step, \algo-0 skips recomputation entirely, incurring zero runtime overhead.

Using the synchronous workload, we draw two key insights from the results in Figure~\ref{fig-eval-e2e} and Figure~\ref{fig-eval-zero}. First, \algo-0, despite its minimal link-time cost, \algo-0 preserves accuracy remarkably well. Second, the ``attention sink'' phenomenon disappears in the middle (Figure~\ref{fig-eval-zero}), reinforcing the importance of mitigating chunk-initial tokens' influence on subsequent attention computations. On the other hand, the previously raised concern in \baseline\ regarding limited cross-attention across chunks proves less impactful in practice. Query and decoded tokens can still attend to all earlier chunks, enabling effective aggregation of cross-chunk information.

\textbf{Discussion of lengthy outputs in sparsity algorithms.} \algo\ variants occasionally show reduced accuracy on cases such as (MultiNews, Llama 3.1) and (Needle, Yi). However, this drop stems not from incorrect answers but from unnecessarily lengthy outputs. For the example in Figure~\ref{fig-algo}, \algo-0 correctly begins with ``Chrysan Company'' but continues with unrelated content such as ``and that Derek is living in ...,'' which lowers F1 or ROUGE-L scores. We observe similar behaviors in other sparsity-based algorithms such as StreamingLLM~\cite{xiao2023sink}, H2O~\cite{zhang2024h2o}, and Quest~\cite{Quest-ICML24}. Such behavior undermines the primary goals of sparsity---reducing latency and resource usage. We leave a more detailed investigation of this behavior to future work.

\subsection{Latency and Throughput of \system}

\begin{figure}[t]
    \centering    \includegraphics[width=\linewidth]{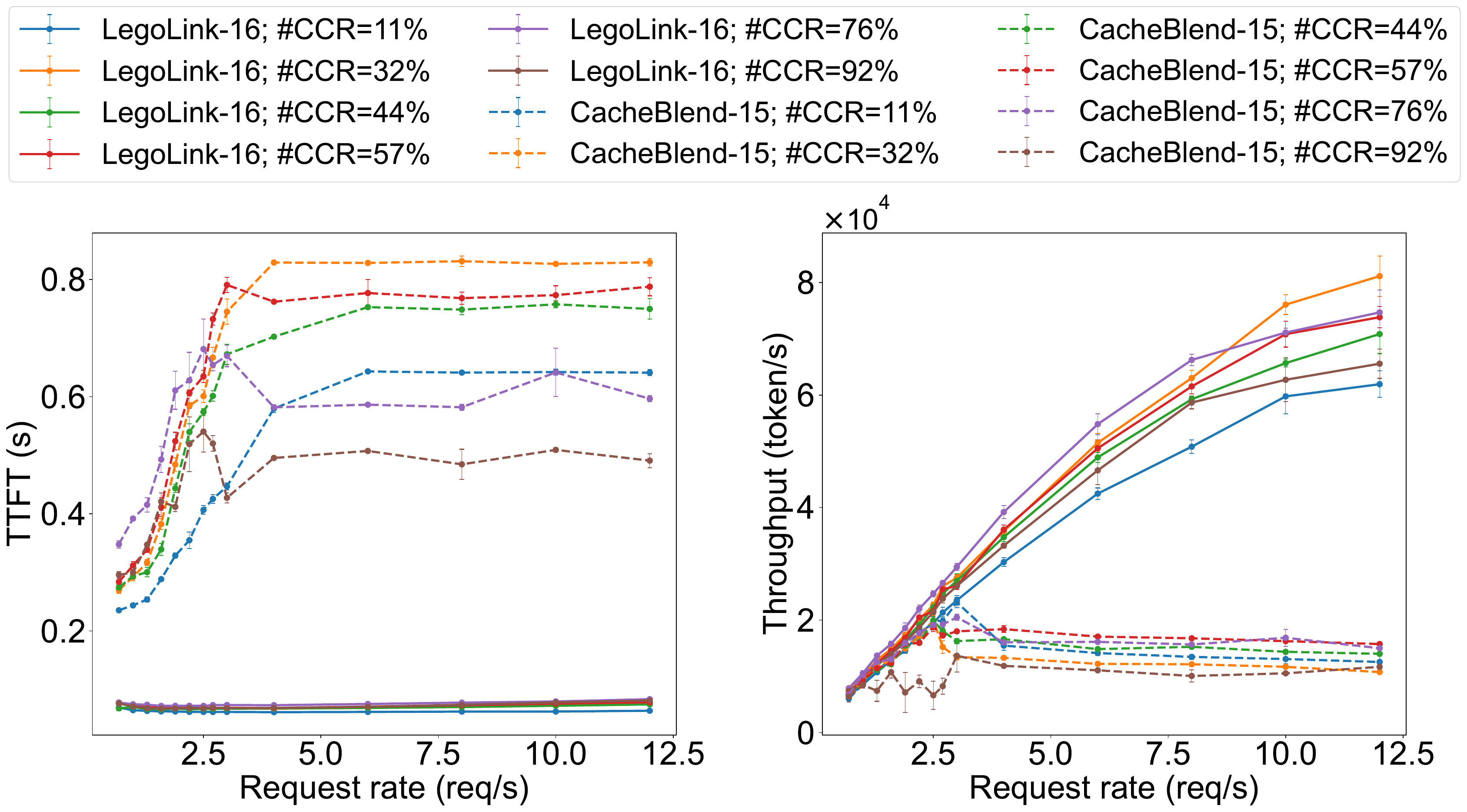}
    \caption{Latency and throughput comparison of \algo-16 and \baseline-15 under asynchronous workloads with varying request rates and context cache ratios (CCR). Each data point represents the average and standard deviation from five experiments. \algo-16 is shown using solid lines, while \baseline-15 is represented with dashed lines. Two algorithms with the same CCR are shown in the same color.}
    \label{fig-eval-latency-thpt}
\end{figure}

We employ the artificial asynchronous workloads on \algo-16 (16 is the block size of vLLM and a moderate number of tokens to recompute) and \baseline-15, presenting results in Figure~\ref{fig-eval-latency-thpt}. Notably, the numbers in this section should be interpreted cautiously when considering real-world scenarios.

Regarding \textit{TTFT} versus request rates (left of Figure~\ref{fig-eval-latency-thpt}), we observe three key trends. First, \algo-16 achieves up to an $8\times$ reduction in \textit{TTFT} compared to \baseline-15. Second, as the Context Cache Ratio (CCR) increases, \algo-16 remains stable \textit{TTFT}, whereas \baseline-15 fluctuates around 0.5 seconds. This stability likely results from \algo-16 generating fewer intermediate results; higher CCR reduces available memory for intermediate computation, but \algo-16 incurs less recomputation overhead than \baseline-15. Third, \textit{TTFT} plateaus as request rate increases, rather than growing exponentially. This plateau reflects vLLM’s scheduling policy, which limits the number of concurrent running requests based on available memory. If we included the \textit{TTFT} of all waiting (queued) requests, the average \textit{TTFT} would approach infinity.

Regarding throughput versus request rates (right of Figure~\ref{fig-eval-latency-thpt}), we observe two notable facts. First, \algo-16 achieves a throughput that is up to $7\times$ higher than \baseline-15, as it recomputes fewer tokens, allowing more requests (about $7\times$) to be processed simultaneously. Second, as CCR increases, \algo-16's throughput continues to improve until the CCR reaches a threshold (approximately $30\%$), beyond which further increases in CCR lead to a reverse effect because requests start to severely interfere with each other. In contrast, \baseline-15's throughput remains constant as it becomes incapable of handling additional requests. 

\subsection{\system's Performance on Long Context}

\begin{figure}[t]
    \centering
    \includegraphics[width=0.7\linewidth]{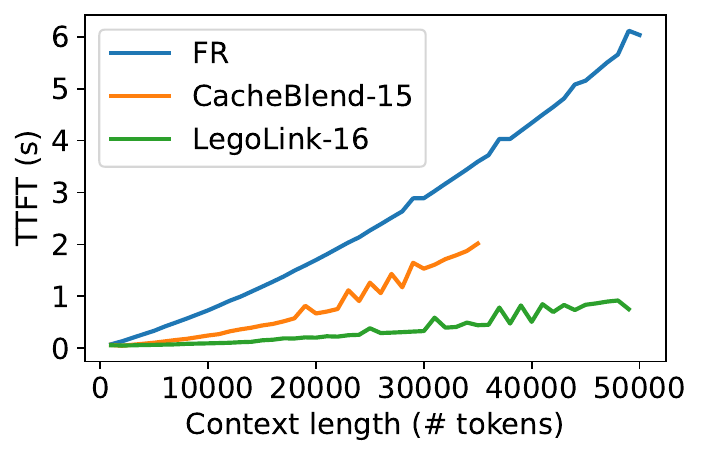}
    \caption{TTFT vs. context length of \textit{FR}, \baseline-15, and \algo-16, using a fixed chunk size of 512 tokens. For \textit{FR}, we do not compile context cache to display its full quadratic time complexity trend, as it would otherwise run out of memory earlier than \baseline-15 and \algo-16.}
    \label{fig-eval-long-context}
\end{figure}

We send requests of varying context lengths with a fixed chunk size (512 tokens) synchronously to \system, yielding two observations from the results (Figure~\ref{fig-eval-long-context}). First, as context length increases, the \textit{TTFT} of both \textit{FR} and \baseline-15 grows quadratically, while \algo-16 exhibits nearly linear growth. This difference arises because \textit{FR} and \baseline-15 have time and resource complexities of $O(N^2)$, while \algo-16 operates with a complexity of $O(kN)$, where $k$ represents the number of recomputed tokens ($k \ll N$). Second, \algo-16 supports a longer context length compared to \baseline-15. Specifically, \baseline-15 encounters an out-of-memory (OOM) error at approximately 35,000 tokens, while \algo-16 avoids OOM until the context length reaches 50,000 tokens. This difference is due to \baseline-15's need to recompute more tokens and generate additional intermediate results, leading to higher GPU memory usage.

\section{Related Work}
\label{sec-related}

This work formalizes Position-Independent Context Caching (PIC) and advances the state of the art in this emerging area. Below, we outline the broader design space relevant to our work.

\textbf{LLM-serving optimizations.}
Numerous systems have recently emerged to improve LLM serving efficiency. vLLM~\cite{kwon2023vllm} introduces PagedAttention to achieve high throughput, while SGLang~\cite{zheng2023sglang} provides both a domain-specific frontend language and an optimized backend runtime. DeepFlow~\cite{hu2025deepflow} integrates the advantages of existing research work into a system running on Ascend accelerators at Huawei Cloud. In addition to full systems, researchers have proposed scheduling techniques such as disaggregated prefill and decode~\cite{zhong2024distserve, hu2024memserve, tetriinfer-2024, patel2024splitwise}, continuous batching~\cite{yu2022orca}, and multi-LoRA integration~\cite{sheng2023s, li2024caraserve}. Storage-related optimizations such as KV-cache-centric inference systems~\cite{qin2407mooncake, hu2024memserve} also contribute to this space.

\balance

\textbf{Context Caching (CC).} Two primary types of context caching have emerged. First, prefix-based CC emerged in late 2023, represented by Pensieve~\cite{yu2023stateful-pensieve}, CacheGen~\cite{liu2023cachegen}, and SGLang~\cite{zheng2023sglang}. Recently, vendors such as Kimi~\cite{kimi-context-caching} and Gemini~\cite{gemini-context-caching} have begun offering explicit CC APIs. Second, PIC emerged in mid-2024 and \baseline~\cite{yao2024cacheblend} represents the first attempt to tackle the PIC challenge, although it does not formally define the challenge. PromptCache~\cite{gim2024prompt} aims to support PIC, but its reuse remains position-dependent. In this paper, we formally define PIC and advance the state of the art by introducing \algo, a low-overhead or even zero-overhead linking algorithm.

\textbf{Sparsity.} Sparsity plays a crucial role in improving long-context inference and falls into two types: dynamic and static. First, dynamic sparsity (e.g., H2O~\cite{zhang2024h2o}, Quest~\cite{Quest-ICML24}, ArkVale~\cite{chen2024arkvale}, RaaS~\cite{hu2025raas}) determines important tokens at runtime. Second, static sparsity (e.g., Longformer~\cite{longformer}, StreamingLLM~\cite{xiao2023sink}) relies on predefined sparse patterns. \baseline\ leverages dynamic sparsity while \algo\ leverages static sparsity to enable efficient linking.

\textbf{Retrieval-Augmented Generation (RAG).} RAG~\cite{li2022survey,jin2024ragcache,gao2023retrieval,jeong2024adaptive,ram2023context,mao2020generation} enhances LLMs' capabilities by integrating external knowledge to improve factuality and relevance. For example, at the application level, Adaptive-RAG~\cite{jeong2024adaptive} dynamically selects retrieval and generation strategies based on query complexity. At the system level, RAGCache~\cite{jin2024ragcache} reduces latency by caching and reusing intermediate states from retrieved documents. PIC has strong potential in RAG scenarios, where reusing documents' KV cache across requests can yield significant performance gains.

 % We have balance in here
\section{Conclusion}
\label{sec-conclusion}

In this paper, we have formalized the Positional-Independent-Cache (PIC) framework. Within this framework, we have proposed \system, a system that incorporates the \algo\ algorithm to address key limitations of existing approaches. By leveraging static attention sparsity, \algo\ significantly reduces recomputation complexity in the link step while maintaining accuracy. Extensive evaluation across six datasets and three LLM models has shown that \system\ achieves significant improvements in \textit{TTFT} and throughput compared to existing systems, with minimal or no accuracy loss.

\section*{Acknowledgments}

This work was partially supported by National Natural Science Foundation of China under Grant No.92464301. We would also like to thank the anonymous reviewers for their insightful comments and suggestions, which helped improve the quality of this paper. 

% \textbf{Do not} include acknowledgements in the initial version of
% the paper submitted for blind review. 

% If a paper is accepted, the final camera-ready version can (and
% usually should) include acknowledgements.  Such acknowledgements 
% should be placed at the end of the section, in an unnumbered section
% that does not count towards the paper page limit. Typically, this will 
% include thanks to reviewers who gave useful comments, to colleagues 
% who contributed to the ideas, and to funding agencies and corporate 
% sponsors that provided financial support.

\section*{Impact Statement}

This paper presents work whose goal is to advance the field of Machine Learning. There are many potential societal consequences of our work, none of which we feel must be specifically highlighted here.

\newpage
\balance
\bibliography{example_paper}
\bibliographystyle{icml2025}

\newpage
\appendix
% \onecolumn

\section{Runtime Overhead of \baseline}

\begin{figure}[t]
    \centering
    \includegraphics[width=1\linewidth]{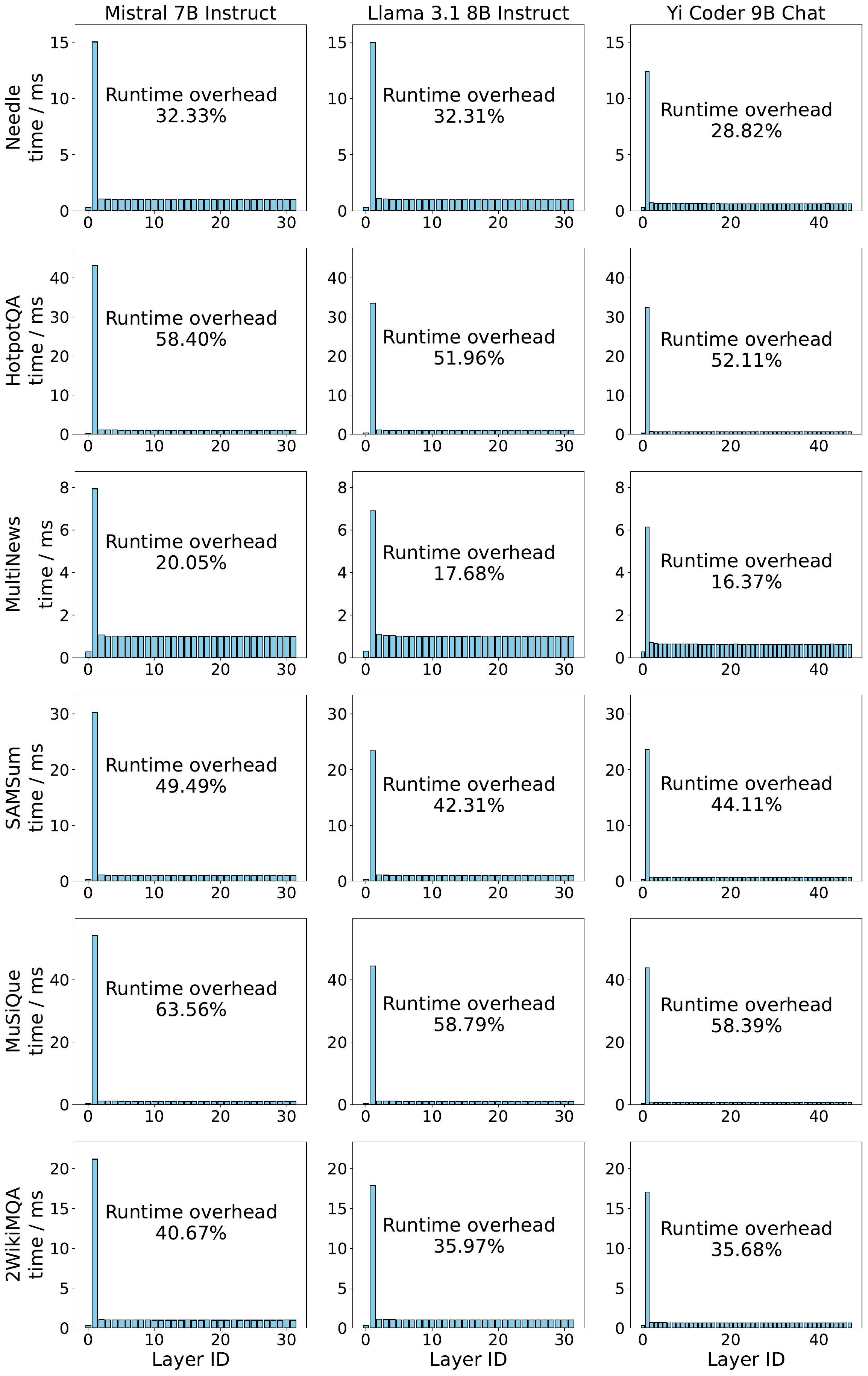}
    \caption{TTFT breakdown of \baseline-15.}
    \label{fig-eval-breakdown}
    % \vskip -0.2in
\end{figure}

We evaluate the runtime overhead of \baseline-15, using the synchronous workloads described in Section~\ref{sec-eval}. Figure~\ref{fig-eval-breakdown} presents the average time-to-first-token (\textit{TTFT}) across 200 requests, with a breakdown of computational costs. The second transformer layer---where 15\% of tokens for recomputation are dynamically selected---contributes $16.37\% - 63.56\%$  of the total \textit{TTFT}. This finding underscores the substantial overhead introduced by dynamic sparsity in \baseline. In contrast, static sparsity, which predefines the recomputed tokens, significantly reduces this overhead, as detailed in Section~\ref{sec-algo}.

\balance

\section{Implementation Details}

We implement \texttt{KVCache}\footnote{As building a highly efficient \texttt{KVCache} is not the core focus of this paper, we build a minimal working system.} based on vLLM's original memory management and prefix caching subsystem with the following changes.
First, we add a cache-ID-based indexing mechanism using the sequence group ID as the cache ID. Second, since the original vLLM manages historical KV cache residing in HBM only, we extend it to include DRAM and local filesystem, akin to Mooncake~\cite{qin2407mooncake}. Third, we modify the scheduler to retain block tables and memory for PIC compile requests. Fourth, we also implement helper APIs that allow users to manage the lifecycle of KV cache, such as \texttt{expire\_cache(cache\_id)}. 

We implement \texttt{KVCompile} as a standalone module that handles CC APIs that are similar to those in Kimi~\cite{kimi-context-caching} and Gemini~\cite{gemini}. \texttt{KVCompile} forwards a request to \texttt{Regular Prefill} with maximum generation token set to 0.

We implement \texttt{KVLink} as a parallel module of the \texttt{Regular Prefill}. First, we adapt the model architecture to support masked attention across tokens scattered in different positions. Second, we modify the attention backends to handle data placement, movement, and the computational steps required by PIC algorithms, to ensure efficient recomputation.

\end{document}